\documentclass[11pt]{article}

\usepackage[preprint]{acl}

\usepackage{times}
\usepackage{latexsym}
\usepackage{wrapfig}
\usepackage{float}
\usepackage{comment}
\usepackage{multirow}
\usepackage{url}
\usepackage{xcolor}
\usepackage{rotating}
\usepackage{booktabs}

\usepackage[T1]{fontenc}

\usepackage[utf8]{inputenc}

\usepackage{microtype}

\usepackage{inconsolata}

\usepackage{graphicx}

\newcommand{\ignore}[1]{}
 \newcounter{todocounter}

\title{It's All About the Confidence: An Unsupervised Approach for Multilingual Historical Entity Linking using Large Language Models}

\author{
  \textbf{Cristian Santini\textsuperscript{1,3}},
  \textbf{Marieke van Erp\textsuperscript{2}},
  \textbf{Mehwish Alam\textsuperscript{3}}
\\
  \textsuperscript{1}Department of Humanities, University of Macerata, Italy,
\\
  \textsuperscript{2}KNAW Humanities Cluster, DHLab, Amsterdam, the Netherlands,
\\
  \textsuperscript{3}INFRES Department, Télécom Paris, France,
\\
  \small{
    \textbf{Correspondence:} \href{mailto:c.santini12@unimc.it}{c.santini12@unimc.it}, \href{mailto:marieke.van.erp@dh.huc.knaw.nl}{marieke.van.erp@dh.huc.knaw.nl}, \href{mailto:mehwish.alam@telecom-paris.fr}{mehwish.alam@telecom-paris.fr}, 
  }
}

\begin{document}
\maketitle
\begin{abstract}
Despite the recent advancements in NLP with the advent of Large Language Models (LLMs), Entity Linking (EL) for historical texts remains challenging due to linguistic variation, noisy inputs, and evolving semantic conventions. Existing solutions either require substantial training data or rely on domain-specific rules that limit scalability. In this paper, we present MHEL-LLaMo (\textit{Multilingual Historical Entity Linking with Large Language MOdels}\footnote{This acronym was chosen because its pronunciation resembles the Spanish phrase \textit{``Me llamo''}, which means \textit{``My name is''}, a common way to declare someone's identity.}), an unsupervised ensemble approach combining a Small Language Model (SLM) and an LLM. MHEL-LLaMo leverages a multilingual bi-encoder (BELA) for candidate retrieval and an instruction-tuned LLM for NIL prediction and candidate selection via prompt chaining. Our system uses SLM's confidence scores to discriminate between easy and hard samples, applying an LLM only for hard cases. This strategy reduces computational costs while preventing hallucinations on straightforward cases. We evaluate MHEL-LLaMo on four established benchmarks in six European languages (English, Finnish, French, German, Italian and Swedish) from the 19th and 20th centuries. Results demonstrate that MHEL-LLaMo outperforms state-of-the-art models without requiring fine-tuning, offering a scalable solution for low-resource historical EL. The implementation of MHEL-LLaMo is available on \href{https://github.com/sntcristian/MHEL-LLAMO}{Github}.
\end{abstract}

\section{Introduction}

Entity Linking (EL) is a crucial Information Extraction (IE) task, allowing researchers to automatically trace and link references to named entities to external Knowledge Bases (KBs) for semantic enrichment, document indexing, and linking to external sources. EL has been used in a variety of use cases, such as social network analysis~\citep{groth2024understanding, santini2024art}, media monitoring~\citep{chen2022analyzing}, and artwork and motif recognition~\citep{santini2022knowledge, sierra2024melart}. 

Despite the remarkable performance of EL systems on contemporary benchmarks~\citep{de2022multilingual,plekhanov2023multilingual,limkonchotiwat-etal-2023-mrefined}, this task remains challenging with historical texts due to complex document varieties, noisy inputs, multilinguality, and evolving syntactic and semantic conventions.\footnote{For more details and examples, a comprehensive survey of IE tasks in historical texts has been presented in~\citep{ehrmann2023named}.} Some approaches have been proposed to adapt EL systems to deal with historical documents, relying either on supervised architectures~\citep{linhares2022melhissa} that need training data and careful parameter tuning or rule-based constraints which do not scale well across different domains~\citep{labusch2020named}. More recently, Large Language Models (LLMs) have been used to perform EL on historical texts with zero-shot prompting~\citep{boscariol_evaluation_2024, graciotti-etal-2025-ke}; however, the results obtained are still inferior to those of supervised domain-specific approaches~\citep{labusch2020named, boros2020robust, linhares2022melhissa}.

The goal of this paper is to perform Multilingual Historical Entity Linking (MHEL) in an unsupervised way by using LLMs in combination with a Small Language Model (SLM). Specifically, we use BELA~\citep{plekhanov2023multilingual}, a multilingual XLM-R-based bi-encoder~\citep{conneau-etal-2020-unsupervised} for retrieving a set of candidates from Wikidata similar to a given mention. Then, we leverage \textit{prompt chaining} to guide an LLM in either selecting the most plausible candidate from the set of Wikidata entities or predicting a NIL entity when no candidate corresponds to the entity referenced in the text. Based on the findings of~\citet{ma-etal-2023-large}, we investigate how confidence scores returned by the SLM may improve the performance of LLMs by allowing them to discriminate between easy and hard samples. The main hypothesis of this work is that using LLMs only for hard samples to refine the bi-encoder’s results provides a threefold advantage: 
\begin{enumerate}
    \item The background knowledge of LLMs is leveraged to deal with complex named entity mentions;
    \item By keeping the bi-encoder's results for easy samples, LLMs are prevented from hallucinating on simple cases;
    \item By using LLMs only for hard samples, energy costs for model inference are minimized.
\end{enumerate}

Therefore, the main contribution of this work is to propose an unsupervised ensemble architecture for MHEL called MHEL-LLAMO (\textit{Multilingual Historical Entity Linking with Large Language MOdels}), where an LLM performs sequential NIL prediction and candidate selection over a block of entities returned by a bi-encoder only for hard samples, i.e., when confidence scores of the bi-encoder do not reach a minimum threshold. To prove the effectiveness of our approach, we test both a vanilla combination of BELA with LLMs and our ensemble solution MHEL-LLAMO on several datasets with texts spanning from the 19th to the 20th century written in six languages (English, Finnish, French, German, Italian and Swedish). 

Our empirical findings show that MHEL-LLAMO outperforms most of the specialized models for MHEL and suggest that adaptively using LLMs via prompt chaining to refine the bi-encoder's results leads to robust performance on multilingual historical texts belonging to different genres and represents a novel solution to democratize access to advanced IE techniques for low-resource historical languages. To the best of our knowledge, this is the first successful attempt to use LLMs to adapt general-purpose EL systems to historical texts. Datasets, source code and pre-trained models are publicly available to enhance reproducibility.\footnote{\url{https://github.com/sntcristian/MHEL-LLAMO}.}

\section{Related Work}
\label{sec:related_work}
In this section, we discuss related work on historical IE and EL.  

\begin{figure*}[t]
    \centering
    \includegraphics[width=0.9\textwidth]{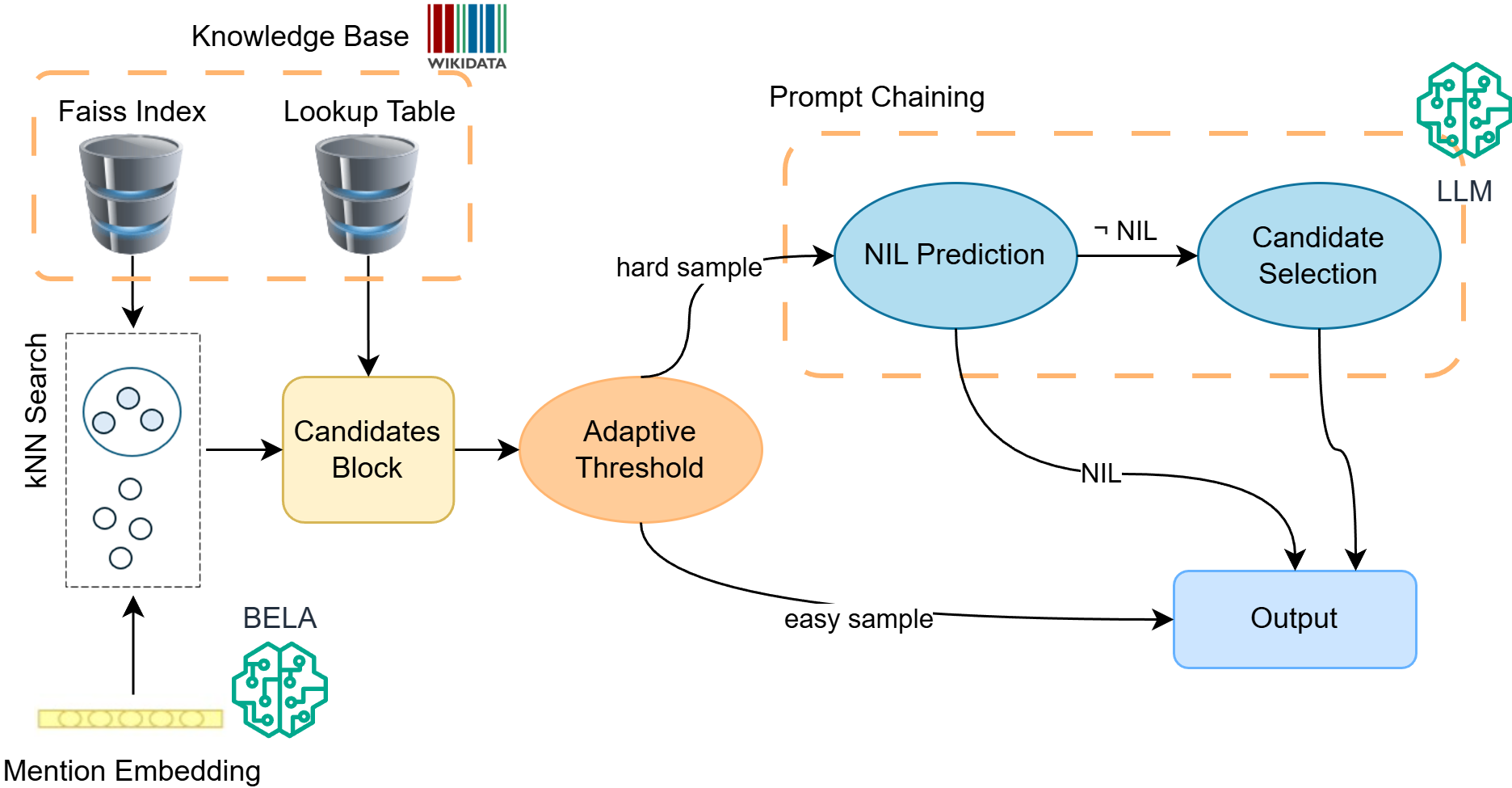}
\caption{Overview of the MHEL-LLaMo architecture. The system combines BELA's bi-encoder for candidate retrieval, a KB consisting of a Faiss Index, and a lookup table for returning similar candidates from Wikidata with metadata, an adaptive threshold to filter easy samples and an instruction-tuned LLM for NIL prediction and candidate selection for hard samples.}
    \label{fig:mhel-llamo}
\end{figure*}

\subsection{Neural Approaches for MHEL}
\label{sec:mhel}

Historical texts present important challenges for EL systems due to the complexity of historical texts, the presence of OCR errors, and the high frequency of NIL entities, i.e., entities that do not appear in a KB. To solve these problems, several specialized MHEL systems have been proposed in recent years. \textit{SBB}~\citep{labusch2020named} is a multilingual framework supporting English, German, and French languages, structured around three core modules: (1) candidate retrieval using word embeddings and $k$-NN search on Wikipedia titles, with temporal filtering based on document publication dates and Wikidata information; (2) candidate evaluation via BERT-based similarity scoring between mention contexts and Wikipedia entity sentences; and (3) candidate re-ranking using Random Forest classification on combined similarity scores. Despite the originality of this work, the proposed approach has potential disadvantages due to the use of Wikidata information for filtering candidates before retrieval, which may lead to a higher false negative rate.

The \textit{L3i} system~\citep{boros2020robust} also presents a neural approach for MHEL on English, French, and German. This method builds upon the end-to-end EL framework of~\citet{kolitsas-etal-2018-end} by adding three key enhancements: an OCR error correction module, a cross-lingual vector index derived from multilingual Wikipedia entity representations, and post-processing heuristics that rerank candidates and identify NIL entities using DBpedia information. Building on this foundation,~\citet{linhares2022melhissa} introduced MELHISSA, an enhanced version of the L3i system also applied to Swedish and Finnish. MELHISSA extends L3i by incorporating 18 specialized post-processing filters for candidate reranking and NIL entity prediction, utilizing string distance metrics such as Levenshtein distance alongside information from both Wikidata and DBpedia. Despite MELHISSA and L3i's robust performance across the HIPE-2020~\citep{ehrmann2020introducing} and NewsEye~\citep{hamdi2021multilingual} datasets, their extensive reliance on rule-based parameters and training data creates reproducibility challenges in the absence of large annotated datasets and extensive empirical tests.

To overcome the limitations of the approaches herein described, this work proposes to leverage the background knowledge of LLMs to perform NIL prediction and candidate selection in MHEL in an unsupervised way. While these models notably suffer from hallucinations, which may limit their efficiency in retrieving the correct candidate from Wikidata with a zero-shot approach, our idea is to adopt a pre-trained multilingual bi-encoder model for candidate retrieval. Further, we frame the NIL prediction and candidate selection tasks as multiple-choice question-answering tasks for LLMs, an activity for which instruction-tuned agents are thoroughly trained. 

\subsection{Historical IE with LLMs}
\label{sec:hist_llm}

Since the advent of LLMs, numerous approaches of these models for IE tasks on historical texts have been investigated. In~\citet{gonzalez-gallardo_yes_2023, gonzalez-gallardo_leveraging_2024}, ChatGPT~\citep{brown2020language}, and open-source LLMs (including LLaMa2 \& 3, Mistral and Zephyr) were used in zero-shot prompting for Named Entity Recognition (NER) on the historical corpus HIPE-2022~\citep{ehrmann2022overview}. While the experiments show that LLMs are generally capable of understanding IE instructions and recognizing simple classes of entities (e.g., person and location), they present notable issues due to the necessity of adopting post-processing strategies to control output formatting and remove instances of new classes of entities not present in the benchmark. 

Recently,~\citet{santini_named_2024} compared open-source LLMs with models based on the GliNER~\citep{zaratiana-etal-2024-gliner} architecture for NER on a corpus of 19th-century Italian philological texts. The main insight of this research was to show that LLMs exhibit important biases in the recognition of named entities, underperforming on domain-specific classes such as bibliographic references. Moreover,~\citet{graciotti-etal-2025-ke} presented a novel benchmark study on multilingual historical music periodicals where they tested GPT-4o mini and LLaMa3-70B in NER and EL with zero-shot prompting. While the results show how large instruction-tuned models surpass 50\% of accuracy in both tasks using zero-shot prompting, the study also emphasizes that these models perform poorly on long-tail entities, i.e., entities which occur less frequently in web-crawled corpora. To the best of our knowledge, MHEL-LLaMo is the first successful attempt to apply LLMs for MHEL.

\section{Methodology}
\label{sec:methodology}

MHEL-LLaMo is based on three main components:
\begin{enumerate}
    \item A candidate retrieval module based on BELA~\citep{plekhanov2023multilingual};
    \item A lookup table containing structured metadata about Wikidata entities, stored in an SQLite database; 
    \item An instruction-tuned LLM which performs NIL prediction and candidate selection based on a textual mention of an entity and a set of similar Wikidata candidates returned by the retrieval module. 
\end{enumerate}

A high-level representation of the approach is presented in Figure~\ref{fig:mhel-llamo}. 

\subsection{Candidate Retrieval}

For candidate retrieval, BELA's XLM-R bi-encoder is used to encode an entity mention and its surrounding context in a mention embedding~\citep{conneau-etal-2020-unsupervised}. This embedding is used to perform $k$-NN search on a dense vector index~\citep{douze2024faiss} of the same dimensionality, containing latent representations of Wikidata entities extracted from Wikipedia in 97 languages. The $k$-NN search returns the top-$k$ most similar candidates to a mention embedding, ranked by inner product. We selected BELA as the candidate retrieval module since it is currently the only multilingual bi-encoder for entity linking that provides both pre-trained model weights and pre-computed entity embeddings. Alternative multilingual approaches such as mReFinED~\citep{limkonchotiwat-etal-2023-mrefined} or mGENRE~\citep{de2022multilingual} either do not release pre-trained models or do not support candidate retrieval via FAISS indexing, making them unsuitable for our unsupervised pipeline.

The lookup table component is used to enrich the information about every candidate with structured information from Wikidata. For every candidate retrieved by the bi-encoder, the lookup table returns its label and description in the language of the text, along with the earliest date found in Wikidata and its entity type. This information allows the LLM to better discriminate the correct candidate based on KB information. 

\subsection{Adaptive Threshold}

Inspired by the findings of~\citet{ma-etal-2023-large}, we exploit the confidence scores of the retrieval module (i.e., the inner product) to discriminate between easy and hard samples. Following their approach, we perform NIL prediction and candidate selection only if the scores of the candidates are below a certain threshold, thus considering a mention as a hard sample. If at least one candidate has a score higher than or equal to the threshold, we link the mention to the candidate with the highest similarity. 

\subsection{Candidate Selection}

We then use an instruction-tuned LLM via prompt chaining to sequentially perform NIL prediction and candidate selection by evaluating the similarity between a text mention and each Wikidata candidate returned by the bi-encoder. The choice of LLM was guided by two criteria: (1) strong performance on the target languages as reported by the EuroEval benchmark\footnote{\href{https://euroeval.com/}{https://euroeval.com/}}, and (2) interoperability with the Hugging Face Transformers library to ensure interoperability. Based on these criteria, we selected Mistral-Small-24B-Instruct\footnote{\href{https://huggingface.co/mistralai/Mistral-Small-24B-Instruct-2501}{mistralai/Mistral-Small-24B-Instruct-2501}} for texts in English, French, German, and Italian, Gemma-3-27B-it\footnote{\href{https://huggingface.co/google/gemma-3-27b-it}{google/gemma-3-27b-it}} for texts in Swedish, and Poro-2-8B-Instruct\footnote{\href{https://huggingface.co/LumiOpen/Llama-Poro-2-8B-Instruct}{LumiOpen/Llama-Poro-2-8B-Instruct}} for texts in Finnish.

Our prompt chaining strategy for NIL prediction and candidate selection follows a two-step approach. First, we ask the instruction-tuned LLM to predict whether an entity is NIL by determining whether there is at least one exact match between a text mention and one of the candidates returned by the bi-encoder. If the answer is ``\emph{no}’’, we classify the entity as NIL; otherwise, we run a separate prompt to ask the LLM to select the most similar entity among the candidates with respect to the mention. It is important to note that in this selection step the LLM can still return an empty set if it is not able to confidently determine which is the most plausible entity. If the LLM returns one entity from the candidate set, we consider it as a correct link; otherwise, we classify the entity as NIL. A detailed explanation of the prompts used is available in Appendix~\ref{sec:prompts}. 

\subsection{Hyperparameter Configuration}

This approach does not need fine-tuning since it relies on the frozen weights of BELA and the instruction-tuned LLMs. The only two hyper-parameters that must be specified are the number of candidates for each mention and the threshold for filtering hard samples, which we selected empirically by finding the best configuration on the development set of each dataset. The values used in our experiments for $k$-NN search and the thresholds applied to BELA's inner products are reported in Appendix~\ref{sec:hyper-parameters}.

\begin{table*}[t]
\begin{tabular}{lllllll}
\hline
\textbf{Dataset}& \textbf{Documents}& \textbf{Mentions}& \textbf{Languages}& \textbf{Period}& \textbf{Genre}& \textbf{\% NIL}\\ \hline
HIPE-2020 & 138              & 3,196              & de, en, fr     & 19C-20C & newspapers         & 25.97\\ 
NewsEye   & 93               & 6,226              & de, fi, fr, sv & 19C-20C & newspapers         & 43.4\\ 
AJMC      & 44               & 1,090              & de, en, fr     & 19C     & classical commentaries & 2.41\\
MHERCL    & 1,408 (sent.) & 4,801              & en, it         & 19C-20C & music periodicals      & 29     \\ \hline
\end{tabular}
\caption{Statistics of evaluation datasets including number of documents, mentions, languages, time period, genre, and percentage of NIL mentions.}
\label{tab:datasets}
\end{table*}

\newpage


\begin{table*}[]
\centering
\resizebox{\textwidth}{!}{
\begin{tabular}{llllllll}
\hline
\textbf{Dataset} & \textbf{Approach} & \multicolumn{6}{c}{\textbf{$F_1$}} \\ \hline
& & \textbf{de} & \textbf{en} & \textbf{fi} & \textbf{fr} & \textbf{it} & \textbf{sv} \\ \hline
\multirow{8}{*}{HIPE-2020} 
& MELHISSA & 0.573 & 0.597 & - & 0.630 & - & - \\
& SBB & 0.506 & 0.393 & - & 0.596 & - & - \\
& L3i & 0.481 & 0.546 & - & 0.602 & - & - \\
& BELA & 0.55 & 0.418 & - & 0.58 & - & - \\
& MHEL-LLaMo$_{van}$ (single) & 0.615 & 0.672 & - & \textbf{0.692} & - & - \\
& MHEL-LLaMo$_{van}$ (chain) & 0.587 & \textbf{0.723} & - & 0.679 & - & - \\
& MHEL-LLaMo$_{\theta}$ (single) & \underline{0.618} & 0.615 & - & 0.672 & - & - \\
& MHEL-LLaMo$_{\theta}$ (chain) & \textbf{0.62} & \underline{0.686} & - & \underline{0.687} & - & - \\ \hline
\multirow{7}{*}{NewsEye}
& MELHISSA & \underline{0.547} & - & \textbf{0.652} & 0.542 & - & \textbf{0.599} \\
& SBB & 0.431 & - & - & 0.444 & - & - \\
& BELA & 0.3 & - & 0.3 & 0.422 & - & 0.41 \\
& MHEL-LLaMo$_{van}$ (single) & 0.488 & - & 0.47 & 0.593 & - & 0.504 \\
& MHEL-LLaMo$_{van}$ (chain) & \textbf{0.556} & - & \underline{0.509} & \underline{0.647} & - & 0.518 \\
& MHEL-LLaMo$_{\theta}$ (single) & 0.444 & - & 0.453 & 0.584 & - & 0.504 \\
& MHEL-LLaMo$_{\theta}$ (chain) & \textbf{0.556} & - & 0.479 & \textbf{0.662} & - & \underline{0.521} \\ \hline
\multirow{6}{*}{AJMC}
& SBB & 0.503 & 0.381 & - & 0.47 & - & - \\
& BELA & 0.443 & 0.338 & - & 0.437 & - & - \\
& MHEL-LLaMo$_{van}$ (single) & \textbf{0.521} & \textbf{0.496} & - & \textbf{0.635} & - & - \\
& MHEL-LLaMo$_{van}$ (chain) & \underline{0.485} & 0.47 & - & 0.57 & - & - \\
& MHEL-LLaMo$_{\theta}$ (single) & \textbf{0.521} & \underline{0.476} & - & \underline{0.6} & - & - \\
& MHEL-LLaMo$_{\theta}$ (chain) & 0.479 & 0.463 & - & 0.57 & - & - \\ \hline
\multirow{8}{*}{MHERCL}
& mGENRE & - & 0.47 & - & - & 0.37 & - \\
& GPT-4o mini & - & 0.6 & - & - & 0.51 & - \\
& LLAMA 3.3 70B & - & 0.61 & - & - & 0.48 & - \\
& BELA & - & 0.47 & - & - & 0.49 & - \\
& MHEL-LLaMo$_{van}$ (single) & - & \textbf{0.7} & - & - & \underline{0.685} & - \\
& MHEL-LLaMo$_{van}$ (chain) & - & \textbf{0.7} & - & - & \textbf{0.698} & - \\
& MHEL-LLaMo$_{\theta}$ (single) & - & 0.65 & - & - & 0.627 & - \\
& MHEL-LLaMo$_{\theta}$ (chain) & - & \underline{0.67} & - & - & 0.629 & - \\ \hline
\end{tabular}
}
\caption{$F_1$ performance comparison across four multilingual entity linking datasets. Best results are shown in \textbf{bold}, second-best results are \underline{underlined}. Our MHEL-LLaMo variants consistently outperform baseline approaches, with particularly strong improvements on HIPE-2020, AJMC and MHERCL datasets.}
\label{tab:results}
\end{table*}

\section{Experimental Setup}
\label{sec:eval_setup}

\subsection{Evaluation Datasets}
We evaluated our approach using four established benchmarks: HIPE-2020~\citep{ehrmann2020introducing}, NewsEye~\citep{hamdi2021multilingual}, AJMC\footnote{HIPE-2020, NewsEye and AJMC are available on the Github repository of the HIPE-2022 Shared Task: \href{https://github.com/hipe-eval/HIPE-2022-data/tree/v2.1-test-all-unmasked}{github.com/hipe-eval/HIPE-2022-data}}~\citep{romanello2024named}, and MHERCL\footnote{\href{https://github.com/polifonia-project/KE-MHISTO/tree/main}{github.com/polifonia-project/KE-MHISTO}}~\citep{graciotti_musical_2025, graciotti-etal-2025-ke}. Overall, these benchmarks include texts in 6 languages: English, Finnish, French, German, Italian and Swedish. They belong to various genres: newspapers, music periodicals and classical commentaries (i.e. scholarly notes to Greek classical works) and span from the nineteenth century (19C) to the twentieth century (20C). Statistics regarding number of documents and number of mentions in each test set and other characteristics are reported in Table~\ref{tab:datasets}. As shown in the table, almost all datasets (except AJMC) have a percentage of NIL mentions higher than 25\%. 

\subsection{Architecture Configuration}

We evaluated MHEL-LLAMO using two distinct configurations:
\begin{itemize}
    \item \textbf{MHEL-LLAMO$_{van}$} (Vanilla Approach): This configuration uses the LLM for NIL prediction and candidate selection for every block of entities returned by BELA;
    \item \textbf{MHEL-LLAMO$_{\theta}$} (Ensemble Architecture): This approach leverages an adaptive threshold based on BELA's confidence scores to distinguish between easy and hard samples. Easy samples are those with high-confidence candidates, while hard samples require LLM intervention for NIL prediction and candidate selection.
\end{itemize}

Each configuration was tested under two settings: (a) \textit{chain}: the LLM uses in sequence the NIL prediction prompt followed by the candidate selection prompt (see Section~\ref{sec:methodology}); (b) \textit{single}: the LLM performs candidate selection and eventually predicts NIL using only the candidate selection prompt. This experimental design yields four distinct variants of MHEL-LLaMo, allowing us to analyze the individual and combined effects of the adaptive threshold mechanism and the prompt chaining strategy.

We determined the optimal hyper-parameters for candidate block size and threshold values by selecting configurations that achieved the best performance on each dataset's development set. These hyper-parameters are detailed in Appendix~\ref{sec:hyper-parameters}, Table~\ref{tab:hyper-parameters}. Experiments were run on two NVIDIA L40S GPUs (2×46GB) with a total computational budget of approximately 60 GPU hours across all datasets and languages. To reduce computational costs, we reported results obtained with a single run.

\subsection{Baseline systems}
The results obtained by MHEL-LLaMo are measured using the $F_1$ metric, following~\citet{ehrmann2022overview}, and compared with those of SoTA models on the same datasets reported in~\citet{linhares2022melhissa},~\citet{ehrmann2022overview} and~\citet{graciotti-etal-2025-ke}. For MELHISSA~\citep{linhares2022melhissa} we consider the results obtained by the best model without post-processing filters for a fair comparison. These baselines are described in Section~\ref{sec:related_work}. Results reported for mGENRE~\citep{de2022multilingual}, GPT-4o mini and LLaMa3.3-70B on MHERCL are taken from~\citet{graciotti-etal-2025-ke}. More recent multilingual entity linking approaches such as mReFinED~\citep{limkonchotiwat-etal-2023-mrefined} or FELA~\citep{fela_2025} were not included as baselines since they do not release pre-trained model weights, preventing reproducible evaluation on our benchmark. All models are tested in the entity disambiguation task, where the correct span of a mention and its surrounding context is given to the model to perform EL or NIL prediction.  

\section{Results}
\label{sec:results}

This section presents the experimental results for MHEL-LLaMo across four MHEL datasets: HIPE-2020, NewsEye, AJMC, and MHERCL. Table~\ref{tab:results} shows the $F_1$ scores achieved by our approach compared to existing SoTA methods. Best results are highlighted in bold, while second-best results are underlined.

\paragraph{HIPE-2020 Performance:} Our approach demonstrates strong performance across all three languages (English, German, French). MHEL-LLaMo$_{van}$ (chain) achieves the best performance on English (0.723 $F_1$), while MHEL-LLaMo$_{\theta}$ (chain) performs best on German (0.620 $F_1$) and achieves the best overall results. The vanilla single configuration excels on French (0.692 $F_1$). Notably, all MHEL-LLaMo variants substantially outperform the baseline systems across different languages.

\paragraph{NewsEye Analysis:} The results on NewsEye present a more challenging scenario, where MHEL-LLaMo outperforms all models in French and German, while MELHISSA maintains superior performance on Swedish and Finnish, probably due to better NIL prediction performance. To better understand this phenomenon, we conducted a thorough error analysis reported in Section~\ref{sec:error_analysis} and an evaluation of NIL prediction performances of best model configurations in Appendix~\ref{sec:nil_eval}. Nonetheless, MHEL-LLaMo$_{\theta}$ (chain) achieves the best performance on French (0.662 $F_1$) and German (0.556 $F_1$) and by a large margin outperforms the supervised system SBB.

\paragraph{AJMC Evaluation:} On the AJMC dataset, MHEL-LLaMo demonstrates clear advantages, with the vanilla single configuration achieving the best results across all three languages: English (0.496), German (0.521), and French (0.635). This represents substantial improvements over the SBB and BELA baselines.

\paragraph{MHERCL Results:} The MHERCL dataset showcases the strongest performance gains for our approach. MHEL-LLaMo$_{van}$ variants achieve the highest $F_1$ scores on both English (0.700) and Italian (0.698), significantly outperforming recent strong baselines including GPT-4o mini and LLAMA 3.3 70B, with an average improvement of $27\%$ over these larger models used in zero-shot setting.

\paragraph{Configuration Analysis:} Comparing the different MHEL-LLaMo variants reveals recurring patterns. MHEL-LLaMo$_{\theta}$ generally outperforms MHEL-LLaMo$_{van}$, particularly on HIPE-2020 and NewsEye. However, the adaptive threshold mechanism ($\theta$) demonstrates its effectiveness primarily on texts belonging to the genre of press articles, probably due to the presence of high-frequency entities contained in these texts (popular places and famous personalities) combined with a high ratio of NIL entities. The choice between chain and single prompt strategies appears dataset-dependent, with chaining showing advantages on HIPE-2020, NewsEye, and MHERCL, while the single prompt excels on AJMC where there is a lower ratio of NIL entities.

\section{Discussion}
\label{sec:discussion}

The experimental results demonstrate that MHEL-LLaMo represents an important advancement in MHEL, achieving substantial improvements over SoTA approaches across multiple datasets. This section analyzes the key findings and their implications.

\paragraph{Performance Advantages:} Our results reveal that MHEL-LLaMo variants consistently outperform existing SoTA models by substantial margins across three of four evaluated datasets. On HIPE-2020, our best configurations achieve improvements of up to 21\% in $F_1$ score for English, 8\% for German, and 10\% for French over MELHISSA. Similarly, we observe substantial performance gains on AJMC, where MHEL-LLaMo achieves relative improvements of 30\% on English, 4\% on German and 35\% on French over the SBB baseline. Most notably, on MHERCL, our approach demonstrates a remarkable average improvement of 27\% over larger models (GPT-4o mini and LLAMA 3.3 70B), achieving $F_1$ scores of 0.700 for English and 0.698 for Italian. These results underscore the effectiveness of combining BELA's efficient candidate retrieval with the LLM's contextual understanding for refined NIL prediction and candidate selection. The approach proves particularly valuable for low-resource historical languages such as historical Italian, suggesting that the bi-encoder component effectively compensates for the LLM's limited exposure to historical language variants during pre-training. 

\paragraph{Effectiveness of Ensemble Strategy:} The empirical results validate the hypothesis proposed by~\citet{ma-etal-2023-large} regarding ensemble strategies for IE tasks, though with important domain-specific nuances. To statistically justify the use of inner products as a proxy for discriminating between easy and hard samples, we conducted a point-biserial correlation analysis (Appendix~\ref{sec:correlation}) measuring the association between BELA's inner product scores and prediction correctness. All correlations are positive and statistically significant ($p < 0.001$), confirming that higher bi-encoder confidence scores are reliably associated with correct predictions and validating the core assumption underlying our adaptive threshold mechanism.

The adaptive threshold mechanism (MHEL-LLaMo$_\theta$) demonstrates clear advantages on newspaper datasets (HIPE-2020 and NewsEye), where it achieves optimal performance for German on HIPE-2020, and for French and German on NewsEye. Conversely, the vanilla approach proves more effective on classical commentaries (AJMC) and music periodicals (MHERCL), where MHEL-LLaMo$_{van}$ achieves the best results. Notably, this issue reveals that confidence scores alone are not always the best proxy for discriminating easy from hard samples. The effectiveness of the threshold mechanism depends on two factors: NIL entity prevalence and entity popularity. For datasets with low NIL rates (AJMC) or high proportions of long-tail entities (MHERCL), the vanilla approach proves more effective, as threshold filtering may bypass mentions where LLM intervention would correct bi-encoder errors. A detailed analysis of this phenomenon is provided in Appendix \ref{sec:correlation}.

\paragraph{Impact of Prompt Chaining Strategy:} An important finding emerges regarding the relationship between prompt strategy and NIL entity prevalence. The chain configuration consistently outperforms the single prompt approach on datasets with high NIL percentages: HIPE-2020 (25.97\% NIL), NewsEye (43.4\% NIL), and MHERCL (29\% NIL). This pattern strongly suggests that LLMs require explicit, separate prompting to accurately identify NIL entities. When using a single prompt that combines candidate selection and NIL prediction, the model exhibits a bias toward selecting existing candidates rather than predicting NIL, leading to degraded performance on datasets with substantial NIL proportions. 

\section{Error Analysis}
\label{sec:error_analysis}

To gain a better understanding of the accuracy of the proposed approach, we conducted an error analysis by sampling 100 false predictions of the best MHEL-LLaMo variants proportionally across all four datasets. The goal of this analysis was to understand if any pattern of semantic relation occurs between the ground truth annotations and the predicted entities. To estimate this, we classified each error in the sample according to 6 classes: false, exact, close, related, broader and narrower. Statistics related to these error classes for the four datasets are reported in Table~\ref{tab:error_analysis}. A detailed explanation of these error classes, with examples, is reported in Appendix~\ref{sec:categories_error_analysis}.

\begin{table}[t] 
\centering
\resizebox{\columnwidth}{!}{
\begin{tabular}{lrrrrr} \hline 
\textbf{Relation} & \textbf{HIPE-2020} & \textbf{NewsEye} & \textbf{AJMC} & \textbf{MHERCL} & \textbf{Tot.} \\ \hline 
False & 15 & 8 & 16 & 20 & 59 \\ 
Exact & 4 & 8 & 4 & 2 & 18 \\ 
Close & 2 & 2 & 1 & 0 & 5 \\ 
Related & 2 & 6 & 3 & 2 & 13 \\ 
Broader & 0 & 1 & 0 & 0 & 1 \\ 
Narrower & 2 & 0 & 1 & 1 & 4 \\ \hline 
\end{tabular} 
}
\caption{Distribution of semantic relations between predicted and ground truth entities in a sample of 100 false predictions from MHEL-LLaMo.}
\label{tab:error_analysis} 
\end{table}

Our error analysis revealed that a consistent number of predictions (41/100) were either exactly referring to the entity mentioned or had a degree of semantic relation with the ground truth. Specifically, we found that 18 predictions were indeed correct (exact). This is primarily because many mentions, especially in NewsEye, are wrongly annotated as NIL and the entity they refer to was correctly retrieved by the model. This phenomenon can be attributed to KB updates, an issue which has already been investigated for EL datasets~\citep{weichselbraun-etal-2019-improving} and requires data curators to adopt versioning strategies to keep up with KB evolution. Another prominent category of errors consists of predictions that are somehow related to the ground truth. For instance, a reference to the Greek hero Odysseus can be linked by the system to the epic poem Odyssey. Moreover, close semantic relations can also appear when an entity may have multiple aliases in Wikidata and the system predicts a different identifier than the one in the ground truth. This is for example the case of the Greek god ``Apollo'', which in Wikidata has two distinct entries, ``Apollo'' (\href{https://www.wikidata.org/wiki/Q37340}{Q37340}) and ``Phoebus'' (\href{https://www.wikidata.org/wiki/Q3382132}{Q3382132}). For more details on the cases found in our error analysis we refer to Appendix~\ref{sec:categories_error_analysis}. Finally, our error analysis revealed that a prominent source of errors is false NIL predictions, along with entities wrongly linked due to OCR errors in the surface form.

\section{Conclusion}
\label{sec:conclusion}
In this paper, we present MHEL-LLaMo, a novel unsupervised approach for MHEL that is inspired by ensemble techniques to combine a multilingual bi-encoder for candidate retrieval with instruction-tuned LLMs for candidate selection. The results obtained by MHEL-LLaMo on six European languages demonstrate not only better accuracy than SoTA specialized models but also robustness toward NIL entities (frequent in historical documents). The improvements achieved on historical benchmarks without the need for fine-tuning prove that this approach can offer a novel scalable solution for MHEL in low-resource scenarios, contributing to the democratization of NLP applications for various case studies in the humanities. 

Future work will investigate the impact of using historical data to train the candidate retrieval module for enhancing the bi-encoder embeddings. Moreover, supervised algorithms (such as Gradient Boosted Trees) can be used to refine the adaptive threshold mechanism in order to take linguistic as well as geometric information into account and more relieably discriminate easy and hard samples. Additionally, ways to take semantic proximity and perspectivism should be taken into account when evaluating MHEL systems~\citep{aroyo2015truth, valette2024does}. Finally, MHEL-LLaMo will be incorporated into existing LLM-based Relation Extraction pipelines~\citep{santini2024combining,balducci2025beyond} for end-to-end Knowledge Extraction on historical texts.

\section{Limitations}
\label{sec:limitations}

Our experiments revealed two important limitations. First, MHEL-LLaMo exhibits reduced performance on Nordic languages, particularly Finnish (best $F_1$: 0.509) and Swedish (best $F_1$: 0.521). This performance gap likely reflects the inherent linguistic complexities of these languages combined with potentially limited multilingual representation in both the bi-encoder and LLM components. The morphological richness and relatively smaller training data availability for Nordic languages present challenges that our current architecture does not fully address. Our NIL prediction analysis (Appendix~\ref{sec:nil_eval}) provides further insight into this phenomenon: Swedish exhibits notably low recall (0.184) in NIL prediction, indicating that Gemma-3-27B-it tends to conservatively predict entities as non-NIL, resulting in a high number of false negatives where actual NIL entities are incorrectly linked to Wikidata candidates. Finnish shows more balanced but still modest performance with Poro-2-8B-it achieving 0.516 recall. These findings suggest that the multilingual capabilities of current open-source LLMs remain limited for Nordic languages, and that language-specific fine-tuning or larger instruction-tuned models may be necessary to improve performance in these settings.

Second, performance on classical commentaries (AJMC) remains lower compared to other genres, with $F_1$ scores ranging from 0.496 to 0.635. Detailed error analysis reveals that the majority of errors stem from linking mentions to literary works and cultural objects. This finding highlights a critical gap in current EL approaches, which are primarily optimized for person, location, and organization entities, and suggests the need for specialized techniques to handle cultural and literary entities that exhibit different referential patterns and contextual cues. Additionally, the AJMC dataset presents the lowest $F_1$ scores for NIL prediction across all languages (0.129--0.167), as a direct consequence of the low proportion of NIL entities in classical commentaries (2.41\%, see Table~\ref{tab:datasets}). This strong class imbalance renders the explicit NIL prediction step in the chain approach prone to generating excessive false positives, making the single prompt strategy more suitable for domains with fewer NIL entities.

Furthermore, BELA embeddings used are trained on a multilingual Wikipedia dump of 2023~\citep{plekhanov2023multilingual}. While this issue may have an impact on recent texts, it does not significantly affect MHEL. Additionally, our error analysis revealed that gold-standard annotations in EL datasets, especially for NIL entities, may lose validity as Wikidata evolves. Although this issue does not originate from our approach, it may affect the validity of benchmarks for historical IE. Nevertheless, this issue can be tackled by adopting versioning techniques in data curation to explicitly state the specific KB release used for data annotation.

Finally, this paper has not explored the use of parameter-efficient fine-tuning techniques, such as LoRA~\citep{hu2022lora}, to train smaller LLMs for NIL prediction and candidate selection. We consider this a promising research direction to reduce the computational cost of our approach and further increase its accessibility.

\section{Ethical Considerations}
While MHEL-LLaMo demonstrates multiple benefits, several potential risks warrant consideration. First, the reliance on LLMs for hard samples incurs environmental costs through GPU inference, though our adaptive approach mitigates this by processing only difficult cases. Second, both BELA and the instruction-tuned LLMs inherit biases from their training data, which may result in unequal performance across languages. Our experiments confirm this, showing reduced accuracy for underrepresented languages like Finnish and Swedish. Finally, linking entities to Wikidata may perpetuate existing biases or gaps in the KB, particularly for non-Western historical figures and cultural artifacts. 

\section*{Acknowledgments}

This research has been carried out during a visiting period spent by Cristian Santini at Telecom Paris (France), under the supervision of Prof. Mehwish Alam. The visiting has received financial support by the University of Macerata through the CIMI scholarlship. Cristian Santini's PhD scholarship has been funded by the PNRR plan under the Next Generation EU program through decree of the Italian Ministry of University and Research (D.M. 117/2023), with the financial support of Grottini Communication S.R.L. (Porto Recanati, Italy). Marieke van Erp's work is funded by the European Union under grant agreement 101088548 - TRIFECTA. Views and opinions expressed are however those of the author only and do not necessarily reflect those of the European Union or the European Research Council. Neither the European Union nor the granting authority can be held responsible for them.

\newpage
\bibliography{custom}

\appendix
\label{sec:appendix}

\section{NIL Prediction and Candidate Selection Prompts}
\label{sec:prompts}

This appendix presents the two prompts used in MHEL-LLaMo to perform entity disambiguation. The two prompts are used in a prompt chaining approach: first, the NIL prediction prompt is used to verify if the LLM finds a plausible match in the list of candidates; then the candidate selection prompt asks the LLM to pick the likeliest entity. Both prompts includes a task instruction, the input text with the target mention marked by \texttt{[ENT]} tags, contextual information related to the text (date, language, genre) and a JSON block listing candidate entities with associated labels, descriptions, types and temporal information.

\subsection{NIL Prediction Prompt}
\begin{quote}
\textbf{System Prompt:}
You are a highly precise multilingual information extraction system specialized in disambiguating entities within noisy historical texts. Your task is to analyse the text provided by the user and determine if the reference marked by [ENT] tags can be associated or not to one of the candidate Wikidata entities provided in the JSON list. Always respond by saying either ``yes'' or ``no''. Do not generate Python code.

\textbf{User Prompt:}
Read the input text written in \texttt{\{language\}}, published in \texttt{\{document\_date\}} and belonging to the genre of \texttt{\{genre\}}.

Answer if the entity mentioned between the [ENT] tags in the input text corresponds to one of the candidate Wikidata entity provided in the json. Give a simple binary answer.

Input Text: \texttt{\{annotated\_text\}}

Candidates: \texttt{\{candidates\_in\_json\}}
\end{quote}

\subsection{Candidate Selection Prompt}

\begin{quote}
\textbf{System Prompt:}
You are an effective multilingual information extraction system specialized in disambiguating entities within noisy historical texts.
Your task is to analyse the text provided by the user and disambiguate the reference marked by [ENT] tags by selecting a Wikidata entity from a given list of candidates. Always respond by returning a JSON-formatted answer; do not generate Python code.

\textbf{User Prompt:}
Read the input text written in \texttt{\{language\}}, published in \texttt{\{document\_date\}} and belonging to the genre of \texttt{\{genre\}}.

Disambiguate the entity mentioned between the [ENT] tags by selecting the most appropriate Wikidata entity from the list of candidates.

Return the corresponding Wikipedia title and Wikidata ID of the selected entity in a JSON object formatted as follows:
\texttt{
\{
``wikipedia\_title'': ``'', ``wikidata\_id'': ``''
\}
}

Make sure to select both the Wikipedia title and the Wikidata ID from the provided list of candidates. Pay attention that the list of candidates may not include the entity mentioned. If none of the candidates match with high confidence the entity tagged with [ENT], return an empty json.

Input Text: \texttt{\{annotated\_text\}}

Candidates: \texttt{\{candidates\_in\_json\}}
\end{quote}

\section{Architecture Hyper-Parameters}
\label{sec:hyper-parameters}

This section reports the hyper-parameters used in the experiments. Threshold to filter easy samples was selected by computing the median of the scores of BELA for correct predictions on the development set. Instead, block size was estimated by measuring $Recall@K$ at 5 steps with $K = \{10, 20, 30, 40, 50\}$ and by finding the best step when there was no significant increment in $Recall@K$ on the development set, i.e., when $Recall@K_{i+ 10} - Recall@K_i < 0.01$.

\begin{table}[ht]
\centering
\resizebox{\columnwidth}{!}{
\begin{tabular}{llrr}
\hline
\textbf{Dataset}           & \textbf{Lang} & \multicolumn{1}{l}{\textbf{Block Size}} & \multicolumn{1}{l}{\textbf{Threshold}} \\ \hline
\multirow{3}{*}{HIPE-2020} & de            & 30                                      & 21.4                                   \\
                           & en            & 20                                      & 21.24                                  \\
                           & fr            & 20                                      & 21.24                                  \\ \hline
\multirow{4}{*}{NewsEye}   & de            & 30                                      & 21.5\\
                           & fi            & 20                                      & 19.57\\
                           & fr            & 20                                      & 21.35                                  \\
                           & sv            & 20                                      & 20.43\\ \hline
\multirow{3}{*}{AJMC}      & en            & 50                                      & 20.7                                   \\
                           & de            & 50                                      & 21.5                                   \\
                           & fr            & 20                                      & 17.92                                  \\ \hline
\multirow{2}{*}{MHERCL}    & en            & 20                                      & 21.5                                   \\
                           & it            & 20                                      & 21.5                                   \\ \hline
\end{tabular}
}
\caption{Architecture hyper-parameters for MHEL-LLaMo experiments across datasets and languages.}
\label{tab:hyper-parameters}
\end{table}

\section{NIL Prediction Evaluation} \label{sec:nil_eval}

To provide further insights into the performance of MHEL-LLaMo on NIL entity prediction, we report in Table~\ref{tab:nil_eval} the Precision, Recall, and F1 scores for NIL prediction across the best-performing model configurations on each dataset and language split.

\begin{table*}[ht]
\centering
\begin{tabular}{llllllll}
\hline
\textbf{Dataset} & \textbf{Lang} & \textbf{LLM} & \textbf{Prompt} & \textbf{$\theta$} & \textbf{Precision} & \textbf{Recall} & \textbf{F1} \\ \hline
\multirow{3}{*}{HIPE-2020} & de & Mistral-24B & chain & yes & 0.440 & 0.642 & 0.522 \\
 & en & Mistral-24B & chain & no & 0.698 & 0.801 & 0.746 \\
 & fr & Mistral-24B & single & no & 0.693 & 0.624 & 0.657 \\ \hline
\multirow{4}{*}{NewsEye} & de & Mistral-24B & chain & yes & 0.684 & 0.510 & 0.584 \\
 & fi & Poro-2-8B & chain & no & 0.649 & 0.516 & 0.575 \\
 & fr & Mistral-24B & chain & yes & 0.839 & 0.562 & 0.673 \\
 & sv & Gemma-3-27B & chain & no & 0.679 & 0.184 & 0.289 \\ \hline
\multirow{3}{*}{AJMC} & de & Mistral-24B & single & no & 0.103 & 0.429 & 0.167 \\
 & en & Mistral-24B & single & no & 0.100 & 0.222 & 0.138 \\
 & fr & Mistral-24B & single & no & 0.091 & 0.222 & 0.129 \\ \hline
\multirow{2}{*}{MHERCL} & en & Mistral-24B & chain & no & 0.588 & 0.803 & 0.679 \\
 & it & Mistral-24B & chain & no & 0.558 & 0.800 & 0.657 \\ \hline
\end{tabular}
\caption{NIL prediction performance of best model configurations across the four historical datasets. Precision, Recall, and F1 are computed specifically for NIL entity detection.}
\label{tab:nil_eval}
\end{table*}

The results reveal several patterns that complement our main findings. First, the chain prompt strategy consistently yields higher recall for NIL prediction, particularly on HIPE-2020 and MHERCL. This confirms that explicitly separating NIL prediction from candidate selection through prompt chaining helps LLMs avoid the bias toward selecting existing candidates observed with the single prompt approach.

Second, the combination of chain prompting with the adaptive threshold mechanism ($\theta$) proves particularly effective for French, both in HIPE-2020 and NewsEye. In these cases, the threshold filtering reduces false positives by preventing the LLM from unnecessarily processing high-confidence predictions, where the bi-encoder already provides reliable results.

Third, the results shed light on the performance gap observed for Nordic languages discussed in Section~\ref{sec:limitations}. Swedish exhibits a notably low recall (0.184) despite reasonable precision (0.679), suggesting that Gemma-3-27B-it has an intrinsic bias toward predicting entities as non-NIL. This conservative behavior leads to a high number of false negatives, where actual NIL entities are incorrectly linked to Wikidata candidates. Finnish shows more balanced but still modest performance, with Poro-2-8B-it achieving 0.516 recall. These findings suggest that the multilingual capabilities of current open-source LLMs remain limited for low-resource languages, and that language-specific fine-tuning or larger instruction-tuned models may be necessary to improve NIL prediction in these settings.

Finally, the AJMC dataset presents the lowest F1 scores for NIL prediction across all languages, with values ranging from 0.129 to 0.167. This is a direct consequence of the low proportion of NIL entities in classical commentaries (2.41\%, see Table~\ref{tab:datasets}), which creates a strong class imbalance. In such scenarios, the single prompt strategy proves more effective, as the explicit NIL prediction step in the chain approach tends to generate excessive false positives when NIL entities are rare.

\section{Confidence Score Correlation Analysis} \label{sec:correlation}

To provide a statistical justification for using the bi-encoder's confidence scores as a proxy to discriminate between easy and hard samples, we conducted a point-biserial correlation analysis measuring the association between the inner product scores produced by BELA during candidate retrieval and the correctness of the final prediction. Results are reported in Table~\ref{tab:correlation}.

\begin{table}[ht]
\centering
\begin{tabular}{llcc}
\hline
\textbf{Dataset} & \textbf{Lang} & \textbf{$r_{pb}$} & \textbf{p-value} \\ \hline
\multirow{3}{*}{HIPE-2020} & de & 0.421 & $< 10^{-39}$ \\
 & en & 0.389 & $< 10^{-9}$ \\
 & fr & 0.384 & $< 10^{-43}$ \\ \hline
\multirow{4}{*}{NewsEye} & de & 0.335 & $< 10^{-31}$ \\
 & fi & 0.354 & $< 10^{-9}$ \\
 & fr & 0.371 & $< 10^{-42}$ \\
 & sv & 0.455 & $< 10^{-20}$ \\ \hline
\multirow{3}{*}{AJMC} & de & 0.617 & $< 10^{-17}$ \\
 & en & 0.655 & $< 10^{-18}$ \\
 & fr & 0.534 & $< 10^{-14}$ \\ \hline
\multirow{2}{*}{MHERCL} & en & 0.320 & $< 10^{-39}$ \\
 & it & 0.385 & $< 10^{-59}$ \\ \hline
\end{tabular}
\caption{Point-biserial correlation ($r_{pb}$) between BELA confidence scores and prediction correctness across datasets and languages. All correlations are positive and statistically significant.}
\label{tab:correlation}
\end{table}

All correlations are positive and statistically significant ($p < 0.001$), confirming that higher bi-encoder confidence scores are reliably associated with correct predictions. This finding validates the core assumption underlying MHEL-LLaMo's adaptive threshold mechanism: confidence scores provide a meaningful signal for identifying samples that can be resolved without LLM intervention.

The correlation strength varies across datasets: AJMC exhibits the strongest correlations ($r_{pb}$ = 0.53--0.66), HIPE-2020 and NewsEye show moderate correlations ($r_{pb}$ = 0.33--0.46), and MHERCL displays the lowest correlations ($r_{pb}$ = 0.32--0.39). However, correlation strength alone does not fully predict the effectiveness of the ensemble strategy. As shown in Table~\ref{tab:results}, the threshold-based approach (MHEL-LLaMo$_\theta$) outperforms the vanilla approach on HIPE-2020 and NewsEye, while the vanilla approach proves more effective on AJMC and MHERCL despite AJMC exhibiting the highest correlations.

This apparent paradox can be explained by considering two interacting factors: NIL entity prevalence and entity popularity. AJMC contains only 2.41\% NIL entities, meaning that nearly all mentions have a correct candidate in Wikidata. In this scenario, many samples are likely to have a valid candidate with low confidence that the LLM can successfully identify through candidate selection. The high correlation in AJMC reflects that the bi-encoder effectively ranks candidates, but the threshold filtering provides limited advantage when NIL prediction is rarely necessary. Applying the LLM consistently across all samples (vanilla approach) allows it to refine candidate selection without the risk of the threshold incorrectly bypassing mentions that would benefit from LLM intervention.

MHERCL presents a different case: despite a substantial NIL rate (29\%), the vanilla approach still outperforms threshold-based filtering. This is explained by the high proportion of long-tail entities in music periodicals (obscure musicians, composers, music venues) that are underrepresented in the bi-encoder's training data. For such entities, confidence scores become less reliable as discriminators: the bi-encoder may return low confidence even when a correct candidate exists, or conversely, confidently retrieve an incorrect candidate. This interpretation is supported by MHERCL exhibiting the lowest correlations among all datasets. In such scenarios, consistent LLM intervention helps correct bi-encoder errors that threshold filtering would otherwise miss.

Conversely, HIPE-2020 and NewsEye exhibit higher NIL rates (25.97\% and 43.4\%, respectively) combined with a prevalence of popular entities (prominent locations, well-known personalities) that are well-represented in the bi-encoder's training data. In these datasets, the moderate correlations combined with reliable confidence scores for popular entities make the threshold strategy effective: high-confidence samples are likely correct and can bypass the LLM, while low-confidence samples genuinely require LLM intervention for both NIL prediction and candidate disambiguation. The threshold mechanism prevents the LLM from second-guessing correct high-confidence predictions while focusing its capacity on genuinely ambiguous cases.

These findings suggest that the effectiveness of the ensemble strategy depends on the interplay between confidence score reliability, NIL entity prevalence, and entity popularity. The adaptive threshold mechanism is most beneficial when two conditions are met: (1) a substantial proportion of NIL entities exists, and (2) the dataset contains predominantly popular entities for which the bi-encoder produces reliable confidence scores. When either condition is not satisfied — low NIL prevalence (AJMC) or high proportions of long-tail entities (MHERCL) — consistent LLM processing proves more effective.

\section{Semantic Relations in Error Analysis} \label{sec:categories_error_analysis}

This appendix provides detailed definitions and examples of the semantic relation categories used in our error analysis (Section \ref{sec:discussion}). To better understand the nature of false predictions made by the MHEL-LLaMo model, we manually analyzed 100 incorrectly predicted entity links sampled proportionally across the four datasets. Each error was classified according to the semantic relationship between the predicted entity and the ground truth annotation, allowing us to distinguish between completely incorrect predictions and those that exhibit varying degrees of semantic proximity to the correct entity.

The classification scheme consists of six categories that capture different types of relationships between predictions and ground truth entities. Concise definitions of each category are reported below, which are further illustrated through concrete examples in Table \ref{tab:semantic_examples}.

\begin{description}
    \item[\textbf{False}] The predicted entity has no semantic relationship to the ground truth entity; the prediction is completely incorrect.
    \item[\textbf{Exact}] The predicted entity refers to the same real-world entity as the ground truth, but uses a different Wikidata identifier.
    \item[\textbf{Close}] The predicted entity is very similar to the ground truth entity, typically representing the same concept with minor differences (e.g., different time periods or alternative names).
    \item[\textbf{Related}] The predicted entity has a clear semantic connection to the ground truth entity, such as through thematic, conceptual, or associative relationships.
    \item[\textbf{Broader}] The predicted entity represents a more general or superordinate concept that encompasses the ground truth entity.
    \item[\textbf{Narrower}] The predicted entity represents a more specific or subordinate concept that is contained within or is a particular instance of the ground truth entity.
\end{description}

\begin{table*}[ht] 
\centering 
\resizebox{\textwidth}{!}{ 
\begin{tabular}{llll} \hline 
\textbf{Semantic Relation} & \multicolumn{3}{l}{\textbf{Example}} \\ \hline 
& Surface & Ground Truth & Prediction \\ \hline 
False & Électre & Electra (\href{https://www.wikidata.org/wiki/Q733444}{Q733444}) & Électre (\href{https://www.wikidata.org/wiki/Q3587592}{Q3587592}) \\ 
Exact & Sophocles  & Sophocles (\href{https://www.wikidata.org/wiki/Q11950683}{Q11950683}) & Sophocles (\href{https://www.wikidata.org/wiki/Q7235}{Q7235}) \\ 
Close & Epoque & L'Époque (1865-1869) (\href{https://www.wikidata.org/wiki/Q63171281}{Q63171281}) & L'Époque (1937–1949) (\href{https://www.wikidata.org/wiki/Q112206500}{Q112206500}) \\ 
Related & OEdipe & Oedipus at Colonus (\href{https://www.wikidata.org/wiki/Q294001}{Q294001}) & Oedipus (\href{https://www.wikidata.org/wiki/Q130890}{Q130890}) \\ 
Broader & Salford & Salford (\href{https://www.wikidata.org/wiki/Q47952}{Q47952}) & Salford (\href{https://www.wikidata.org/wiki/Q1142118}{Q1142118}) \\ 
Narrower & 11. & Iliad (\href{https://www.wikidata.org/wiki/Q8275}{Q8275}) & Ilias, 11 Book (\href{https://www.wikidata.org/wiki/Q1658482}{Q1658482}) \\ \hline 
\end{tabular} } 
\caption{Examples of semantic relations between ground truth and predicted entities.} 
\label{tab:semantic_examples} 
\end{table*}

\end{document}